\documentclass[letterpaper, 10 pt, conference]{ieeeconf}  % 

\IEEEoverridecommandlockouts % This command is only needed if you want to use the \thanks command
\usepackage[utf8]{inputenc}
\usepackage[T1]{fontenc}
\usepackage{hyperref}
\usepackage{url}
\usepackage{booktabs}
\usepackage{amsfonts}
\usepackage{graphicx}
\usepackage{amsmath} 
\usepackage{subcaption}
\usepackage{color}
\usepackage{algorithm}
\usepackage{algpseudocode}

\renewcommand{\vec}[1]{\boldsymbol{#1}}

\DeclareMathOperator*{\argmax}{arg\,max}
\DeclareMathOperator*{\diag}{diag}

\newcommand{\tran}{^{\mathstrut\scriptscriptstyle\top}} 

\title{\LARGE \bf
Factored Contextual Policy Search with Bayesian Optimization
}

\author{Robert Pinsler$^{*1}$, Peter Karkus$^{*2}$, Andras Kupcsik$^{3,4}$, David Hsu$^2$ and Wee Sun Lee$^2$% <-this % stops a space
\thanks{$^{1}$University of Cambridge, UK $^{2}$National University of Singapore, Singapore $^{3}$Bosch Center for AI, Germany $^{4}$Idiap Research Institute, Switzerland.}
\thanks{* denotes equal contribution. PK conceived the idea and performed initial experiments. RP extended the approach and evaluation, and wrote most of the manuscript. %\newline 
Correspondence to {\tt\small rp586@cam.ac.uk}.}%
\thanks{RP receives funding from iCASE RG89603 supported by Nokia. PK, DH and WSL acknowledge funding from Singapore Ministry of Education AcRF grant MOE2016-T2-2-068. PK is supported by the NUS Graduate School for Integrative Sciences and Engineering Scholarship.}
}

\begin{document}

\maketitle
\thispagestyle{empty}
\pagestyle{empty}

%%%%%%%%%%%%%%%%%%%%%%%%%%%%%%%%%%%%%%%%%%%%%%%%%%%%%%%%%%%%%%%%%%%%%%%%%%%%%%%%
\begin{abstract}

Scarce data is a major challenge to scaling robot learning to truly complex tasks, as we need to generalize locally learned policies over different task contexts. Contextual policy search offers data-efficient learning and generalization by explicitly conditioning the policy on a parametric context space. 
In this paper, we further structure the contextual policy representation. We propose to factor contexts into two components: target contexts that describe the task objectives, e.g. target position for throwing a ball; and environment contexts that characterize the environment, e.g. initial position or mass of the ball. Our key observation is that experience can be directly generalized over target contexts. We show that this can be easily exploited in contextual policy search algorithms. In particular, we apply factorization to a Bayesian optimization approach to contextual policy search both in sampling-based and active learning settings. Our simulation results show faster learning and better generalization in various robotic domains. See our supplementary video: \url{https://youtu.be/MNTbBAOufDY}.

\end{abstract}

%%%%%%%%%%%%%%%%%%%%%%%%%%%%%%%%%%%%%%%%%%%%%%%%%%%%%%%%%%%%%%%%%%%%%%%%%%%%%%%%
\section{INTRODUCTION}

Enabling robots to operate in truly complex domains requires learning policies from a small amount of data and generalizing learned policies over different tasks. Policy search methods with low-dimensional, parametric policy representations enable data-efficient learning of local policies~\cite{deisenroth2013survey}. Contextual policy search~(CPS)~\cite{kupcsik2013data, daniel2012hierarchical} further enables generalization over different task settings by structuring the policy. CPS uses an upper-level policy $\pi(\vec \theta \vert \vec s)$ to select parameters $\vec \theta$ of a lower-level policy given context $\vec s$, where the context $\vec s$ specifies the task. The goal is to learn a policy $\pi(\vec \theta \vert \vec s)$ that maximizes the expected reward $\mathbb{E}[\mathcal{R}_{\vec s, \vec \theta}]$.

We propose to further structure the contextual policy representation by introducing a factorization of the context space. In particular, we factorize a context vector $\vec s$ into two components: (1) \emph{target contexts} $\vec s^t$ that specify task objectives, e.g. for a ball throwing task the target coordinates of the ball, and (2) \emph{environment contexts} $\vec s^e$ that characterize the environment and the system dynamics, e.g. initial position of the ball. Formally, we assume that the expected reward is given by $\mathcal{R}_{\vec s, \vec \theta} = \int p(\vec \tau \vert \vec s^e, \vec \theta) R(\vec s^t, \vec \tau) d\vec \tau$, where $\vec \tau$ is a trajectory with unknown dynamics $p(\vec \tau \vert \vec s^e, \vec \theta)$, and $R(\vec s^t, \vec \tau)$ is the reward function. The key difference between $\vec s^t$ and $\vec s^e$ is that the dynamics only depend on $\vec s^e$. We can exploit this property and re-evaluate prior experience in light of a new target context, leading to improved \mbox{data-efficiency and better generalization}.

\begin{figure}
    \centering
    \includegraphics[width=.35\textwidth]{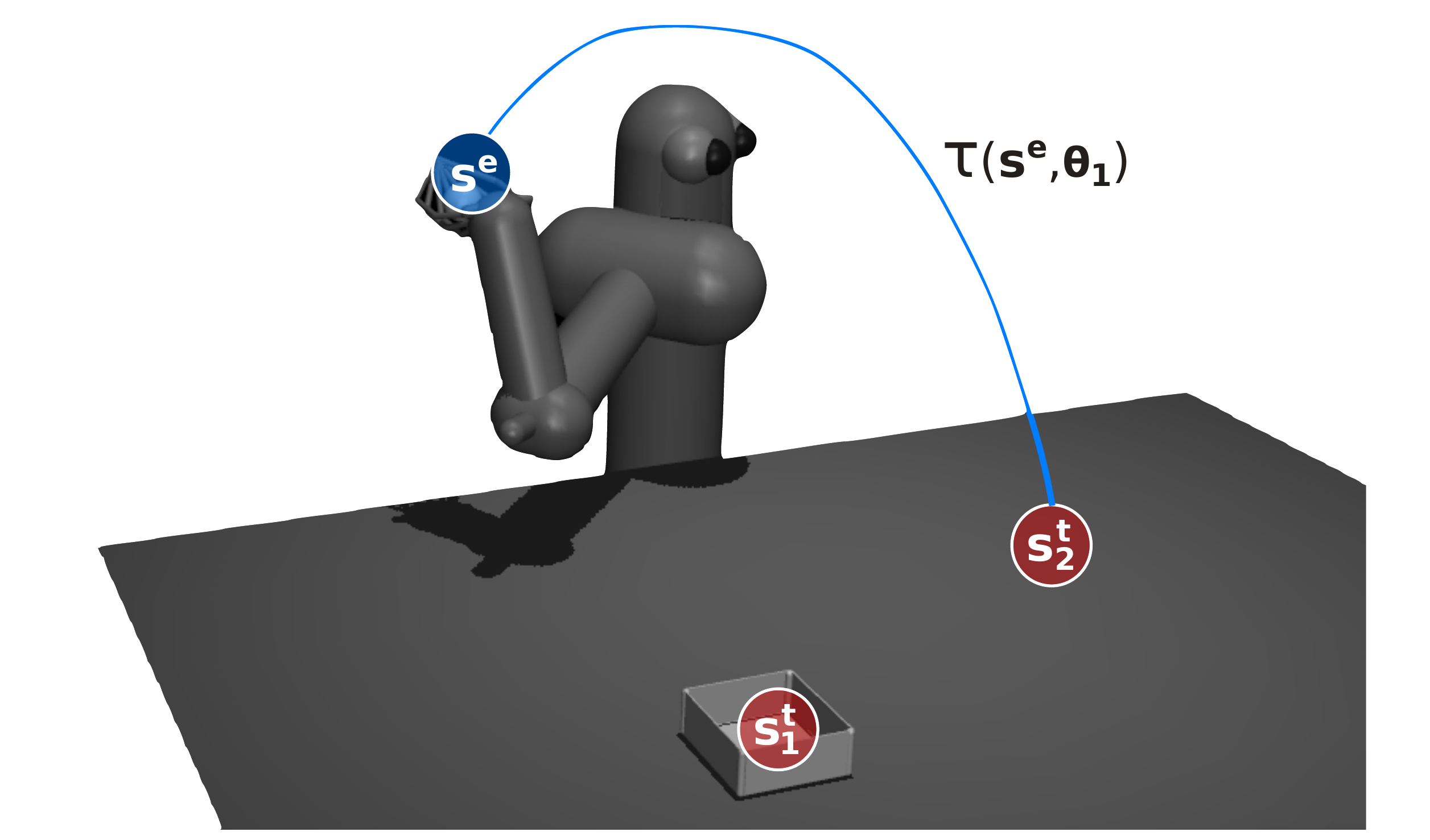}
    \caption{Ball throwing task, where the robot is asked to hit target context $\vec s^t_1$ given initial position $\vec s^e$ of the ball. The robot chooses parameters $\vec \theta_1$ that generates a ball trajectory $\vec \tau \sim p(\vec \tau \vert \vec s^e, \vec \theta_1)$ landing at $\vec s^t_2$. Despite a low reward, knowing that $\vec \tau$ led to $\vec s^t_2$ is beneficial if the robot is asked again to throw near $\vec s^t_2$.
    }
    \label{fig:thrower}
\end{figure}

For example, assume a robot is learning to throw balls at different targets $\vec s^t$ (Fig.~\ref{fig:thrower}). The robot is asked to aim at $\vec s^t_1$. It chooses parameters $\vec \theta_1$, executes the throw, and observes a ball trajectory $\vec \tau_1$ that hits target $\vec s^t_2$, $\vec s^t_2 \neq \vec s^t_1$. This yields a reward $R(\vec s^t_1, \vec \tau_1)$. Assume the robot is now asked to aim at target $\vec s^t_2$. Standard CPS methods try to generalize prior experience solely based on the upper-level policy $\pi(\vec \theta \vert \vec s)$, e.g. by assuming that rewards obtained under similar contexts are correlated. Context factorization instead allows to treat the two context types differently. The target context $\vec s^t_2$ can be used to evaluate $R(\vec s^t_2, \vec \tau_1)$ directly, yielding the exact reward we would get for $\vec \tau_1$ when targeting $\vec s^t_2$. That is because a trajectory is independent from the target context.  The same is not true for environment contexts, and thus we must rely on the upper-level policy to generalize over them.

We demonstrate the benefits of factorization by applying it to CPS approaches based on Bayesian optimization (BO) \cite{brochu2010tutorial}; however, other CPS methods would be also possible. First, we consider a passive learning setting, where the context is given to the robot, and introduce a factored variant of BO for CPS (BO-CPS)~\cite{krause2011contextual, metzen2015bayesian}. We then consider an active learning setting~\cite{fabisch2014active}, where the robot can choose the context during learning. We introduce factored contexts to ACES~\cite{metzen2015active}, a CPS method based on entropy search~\cite{hennig2012entropy}. For moderately low-dimensional search spaces, e.g. when learning pre-structured policies, such global optimization techniques achieve high data-efficiency by directly searching for the optimal parameters using a surrogate reward model. 

So far we assumed that we can re-evaluate the reward function $R(\vec s^t, \vec \tau)$ for arbitrary target contexts.  
This assumption is reasonable in real robot applications, where rewards typically encode objectives defined by the system designer. 
However, if the agent only has access to samples from the reward function, we can still exploit factored contexts by re-evaluating the current trajectory w.r.t. the achieved outcome. This approach can be seen as an extension of hindsight experience replay (HER) \cite{andrychowicz2017hindsight}, a recently proposed data augmentation technique for goal-oriented RL algorithms.

% experiments 
We analyze the proposed methods first on a toy task. We then validate the benefits of context factorization on three simulated robotics environments from the OpenAI Gym~\cite{brockman2016openai}, where we employ dynamic movement primitives \cite{ijspeert2003learning} to efficiently generate trajectories. We show that context factorization is easy to implement, can be broadly applied to CPS problems, and consistently improves data-efficiency and generalization for various robotic tasks. 

\section{RELATED WORK}

There are several CPS approaches that generalize over a context space. One group of work first learns different local policies and then uses supervised learning to interpolate policy parameters over contexts~\cite{da2012learning, metzen2014towards}. These methods are suitable for problems where local policies are available or easy to learn, but they are inefficient otherwise. The second group of work jointly learns local policies and generalizes over the context space~\cite{peters2007applying, kober2012reinforcement, kupcsik2013data}. These approaches were applied to a variety of real-world tasks, including playing table tennis~\cite{kober2012reinforcement}, darts~\cite{kober2012reinforcement} and hockey~\cite{kupcsik2013data}. Although all tasks involve target contexts, generalization over contexts solely relies on correlation. Similarly, CPS approaches based on BO~\cite{metzen2015bayesian, metzen2015active, pinsler2018sample, yang2018learning} learn a probabilistic reward model that generalizes over the context space through correlation. In this paper, we extend two BO approaches with factorization, namely \mbox{BO-CPS}~\cite{krause2011contextual, metzen2015bayesian} and ACES~\cite{metzen2015active}.

To the best of our knowledge, there is no prior work that explicitly factors the context space. Similar ideas are implicitly used by Kober et al. \cite{kober2012reinforcement} who learn a contextual policy for discrete targets while performing a higher-level task. While they map experience gained in one context to another, they do so for estimating discrete outcome probabilities and not for improving the policy. %Hierarchical REPS (HiREPS) \cite{daniel2012hierarchical} frames CPS as a hierarchical RL problem, where a versatile set of options is learned. 
GP-REPS \cite{kupcsik2017model} iteratively learns a transition model of the system using a Gaussian process (GP) \cite{rasmussen2006gaussian}, which is then used to generate trajectories offline for updating the policy. The authors consider generating additional samples for artificial contexts, but they do not define an explicit factorization.

The idea of replacing the goal of a trajectory has recently been explored in HER \cite{andrychowicz2017hindsight}, which increases data-efficiency in goal-based RL tasks with sparse rewards. The key idea is to augment the dataset with additional experience by replacing the original target context of a rollout to be the achieved outcome. Instead of replacing the target context after each rollout, we replace the target context of all previous episodes before each rollout and re-evaluate the entire dataset. Our approach additionally generalizes over environment contexts that are typical in CPS problems. If we have only access to sample rewards, we show how context factorization can be used to extend HER to CPS.

\begin{figure*}[!ht]
    \begin{minipage}[t]{.49\textwidth}
    \vspace{0pt}
    \begin{algorithm}[H]
    \caption{BO-CPS \cite{metzen2015bayesian}}
    \label{alg:bo-cps}
    \begin{algorithmic}
        % \Require Initialize dataset $\mathcal{D} = \{\vec s^e_i, \vec \theta_i, \vec \tau_i \}_{i=1}^N$ randomly \todop{not empty initially?}
        % \Ensure
        \Repeat
            \State Observe $\vec s_q \sim \gamma(\vec s)$
            \\
            \State Learn reward model $p(R  \vert \mathcal{D}, \vec s, \vec \theta)$ from $\mathcal{D}$ (Eq.~\ref{eq:gp-model})
            \State Select $\vec \theta_q \sim \pi(\vec \theta \vert \vec s_q)$ (Eq.~\ref{eq:gp-ucb}, \ref{eq:bocps-policy}) using reward model
            \State Execute rollout $\vec \tau \sim p(\vec \tau \vert \vec s_q, \vec \theta_q)$ with the robot
            \State Add $(\vec s_q, \vec \theta_q , R(\vec s_q, \vec \tau))$ to $\mathcal{D}$
        \Until{Policy $\pi$ converges}
    \end{algorithmic}
    \end{algorithm}
     \end{minipage}
    \begin{minipage}[t]{.49\textwidth}
    \vspace{0pt}  
    \begin{algorithm}[H]
    \caption{BO-FCPS (ours)}
    \label{alg:bo-fcps}
    \begin{algorithmic}
        % \Require Initialize dataset $\mathcal{D} = \{\vec s^e_i, \vec \theta_i, \vec \tau_i \}_{i=1}^N$ randomly \todop{not empty initially?}
        % \Ensure
        \Repeat
            \State Observe $\vec s_q \sim \gamma(\vec s)$, where $\vec s_q = (\vec s^t_q, \vec s^e_q)$
            \State Construct dataset $\mathcal{D}_q$ from $\mathcal{D}$ (Eq.~\ref{eq:map-data})
            \State Learn reward model $p(R_q^t|\mathcal{D}_q, \vec s^e_q, \vec \theta)$ from $\mathcal{D}_q$ (Eq.~\ref{eq:gp-model})
            \State Select $\vec \theta_q \sim \pi(\vec \theta \vert \vec s_q)$ (Eq.~\ref{eq:gp-ucb}, \ref{eq:bocps-policy}) using reward model
            \State Execute rollout $\vec \tau \sim p(\vec \tau \vert \vec s^e_q, \vec \theta_q)$ with the robot
            \State Add $(\vec s^e_q, \vec \theta_q , \vec \tau)$ to $\mathcal{D}$
        \Until{Policy $\pi$ converges}
    \end{algorithmic}
    \end{algorithm}
     \end{minipage}
\end{figure*}

\section{BACKGROUND}

\subsection{Bayesian Optimization for Contextual Policy Search} 

In a CPS problem, the agent observes a \emph{context} $\vec s \sim \gamma(\vec s)$ before each episode, where the context specifies the task setting and $\gamma(\vec s)$ is a distribution over contexts. To solve the task, the agent maintains an upper-level policy $\pi(\vec \theta \vert \vec s)$ over parameters $\vec \theta$ of a lower-level policy, e.g. a dynamic movement primitive~\cite{ijspeert2003learning}. Executing the lower-level policy with parameters $\vec \theta$ generates a trajectory $\vec \tau \sim p(\vec \tau \vert \vec s, \vec \theta)$ that yields reward $R(\vec s, \vec \tau)$. The goal of the agent is to learn an upper-level policy that maximizes the expected reward, 
\begin{equation*}
\mathbb{E}[\mathcal{R}_{\vec s, \vec \theta}] = \iiint \gamma(\vec s)\pi(\vec \theta \vert \vec s)p(\vec \tau \vert \vec s, \vec \theta) R(\vec s, \vec \tau) d\vec \tau d\vec \theta d\vec s.
\end{equation*}
% where $\gamma(\vec s)$ is the context distribution. This distribution does not have to be known in advance and can be estimated from samples.

BO-CPS \cite{krause2011contextual, metzen2015bayesian} frames CPS as a BO problem. BO is a global search method for optimizing real-valued functions, assuming only access to noisy sample evaluations. Starting from a prior belief about the objective, BO employs an acquisition function to guide the sampling procedure. In BO-CPS, a probabilistic reward model $p(R|\mathcal{D}, \vec s, \vec \theta)$ is learned from $N$ data samples $\mathcal{D} = \{\vec s_i, \vec \theta_i, \vec R_i\}_{i=1}^N$, which allows to evaluate potential parameters $\vec \theta$ for a query context $\vec s_q$. BO-CPS commits to a GP prior \cite{rasmussen2006gaussian} with predictive posterior $p(R|\mathcal{D}, \vec s_q, \vec \theta) = \mathcal{N}(\mu_{\vec s_q, \vec \theta}, \sigma^2_{\vec s_q, \vec \theta})$, and uses the GP-UCB acquisition function \cite{srinivas2009gaussian},
\begin{equation} \label{eq:gp-ucb}
    \text{GP-UCB}(\vec s_q, \vec \theta) = \mu_{\vec s_q, \vec \theta} + \kappa \sigma_{\vec s_q, \vec \theta}, 
\end{equation}
where $\kappa$ trades off exploration and exploitation. The policy parameters $\vec \theta$ are selected by the upper-level policy $\pi$, which optimizes the acquisition function given the query context,
\begin{equation} \label{eq:bocps-policy}
    \pi(\vec \theta \vert \vec s_q) = \delta \left(\vec \theta - \vec \theta^* \vert \vec s_q \right),
\end{equation}
where $\vec \theta^* \vert \vec s_q = \argmax_{\vec \theta} \text{GP-UCB}(\vec s_q, \vec \theta)$, and $\delta(\cdot)$ is the Dirac delta function. The algorithm is summarized in Alg.~\ref{alg:bo-cps}.

\subsection{Active Contextual Entropy Search} \label{sec:aces}

Active contextual entropy search~(ACES)~\cite{metzen2015active} is an extension of entropy search~(ES)~\cite{hennig2012entropy} to the active CPS setting, where both the parameters $\vec \theta$ and the context $\vec s$ are chosen by the agent before an episode. ACES maintains a conditional probability distribution $p(\vec \theta^* \vert \mathcal{D}, \vec s) = p(\vec \theta^* = \argmax_{\vec \theta} f(\vec s, \vec \theta) \vert \mathcal{D}, \vec s)$, expressing the belief about $\vec \theta$ being optimal in context $\vec s$. The most informative query point is chosen by maximizing the expected information gain integrated over the context space,
\begin{equation} \label{eq:aces1}
\text{ACES}(\vec s_q, \vec \theta_q) = \sum\nolimits_{c=1}^C G^{\vec s_c}(\vec s_q, \vec \theta_q),
\end{equation}
where $\{\vec s_c\}_{c=1}^C$ is a set of randomly chosen representer points. The expected information gain in context $\vec s_c$ after performing a hypothetical rollout with $(\vec s_q, \vec \theta_q)$ is given by
\begin{equation}
G^{\vec s_c}(\vec s_q, \vec \theta_q) = H[p(\vec \theta^* \vert \mathcal{D}, \vec s_c)] - \mathbb{E}\big[H[p(\vec \theta^* \vert \mathcal{D}^+, \vec s_c)]\big],
\end{equation}
where the expectation is taken over $p(R \vert \mathcal{D}, \vec s_q, \vec \theta_q, \vec s_c)$, and $\mathcal{D}^+ = \mathcal{D} \cup \{\vec s_q, \vec \theta_q, R \}$ is an updated dataset that contains the hypothetical query point. 
%In practice, Eq.~\ref{eq:aces1} is approximated by a sum over the nearest neighbors $\{\vec s\}$ of $\vec s_q$ according to the Mahanalobis distance $d(\vec s_c, \vec s_q) = \sqrt{(\vec s_c - \vec s_q) \smash{\vec \Lambda^{-1}} (\vec s_c - \vec s_q)}$, where $\vec \Lambda$ is a diagonal matrix containing the length-scales of the GP.
In practice, Eq.~\ref{eq:aces1} requires further approximations, which are explained in the original work \cite{hennig2012entropy, metzen2015active}. The algorithm is summarized in Alg.~\ref{alg:aces}.

\subsection{Dynamic Movement Primitives} 

Dynamic movement primitives (DMPs) are often used as lower-level policies in robot learning tasks. A DMP \cite{ijspeert2003learning} is a spring-damper system whose hyper-parameters can be flexibly adapted while retaining the general shape of the movement. These include the final position $\vec y_f$, final velocity $\dot{\vec y}_f$ and temporal scaling $\vec \tau$. The motion is further modulated by a non-linear forcing function $f_{\vec w}(z) = \vec w\tran \Phi(z)$ with basis functions $\Phi(z)$ parameterized by phase variable $z$. The parameters $\vec w$ determine the shape of the movement and can be obtained by imitation learning. Each generated DMP trajectory is followed by the control policy of the robot, which implements a low-level feedback controller.

\section{FACTORED CONTEXTUAL POLICY SEARCH}
In this section, we introduce context factorization and show how it can be integrated into CPS algorithms. 

We propose to factorize a context vector $\vec s$ into two types of contexts, $\vec s = (\vec s^t, \vec s^e)$: 
\begin{itemize}
    \item target contexts $\vec s^t$ which specify the task objective, and
    \item environment contexts $\vec s^e$ which characterize the environment and the system dynamics.
\end{itemize}
Formally, we assume that the reward function is given by $R(\vec s^t, \vec \tau)$\footnote{In general, the reward function may also depend on $\vec s^e$ and $\vec \theta$. We omit this dependence for improved readability; the principle remains the same.}, where the trajectory $\vec \tau$ is generated by unknown system dynamics, $\vec \tau \sim p(\vec \tau \vert \vec s^e, \vec \theta)$. Importantly, while the dynamics function depends on environment contexts, it does not depend on target contexts. This means that we can exchange the target context of a rollout without altering its trajectory, allowing to re-evaluate a rollout under different target contexts. For example, in our ball throwing task~(Fig.~\ref{fig:thrower}) we can re-evaluate a previously observed trajectory pretending we were aiming at a different target. We cannot do the same with environment contexts, e.g. the initial ball pose, because a different initial pose would result in a different trajectory.   
%Note that the same argument does not hold for environment contexts, e.g. a throw with initial ball position $\vec s^e_1$ does not tell anything about the reward for any other position $\vec s^e_2$ apart from the assumed correlation. 
In the following, we exploit factored contexts to reduce the data requirements of CPS algorithms. To what extend factorization can be exploited depends on the knowledge of the reward function $R(\vec s^t, \vec \tau)$. First, we assume that the reward function is fully known or that it can be evaluated for arbitrary targets $\vec s^t$. This allows to construct highly data-efficient algorithms, as we demonstrate on a passive (Section~\ref{sec:bo-fcps}) and active (Section~\ref{sec:faces}) CPS algorithm. In Section~\ref{sec:her}, we drop the assumption of a known reward function and propose an extension of hindsight experience replay \cite{andrychowicz2017hindsight} to the CPS setting using context factorization.

\begin{figure*}[!ht]
    \begin{minipage}[t]{.49\textwidth}
    \vspace{0pt}
    \begin{algorithm}[H]
    \caption{ACES \cite{metzen2015active}}
    \label{alg:aces}
    \begin{algorithmic}
    % \Require Initialize dataset $\mathcal{D} = \{\vec s^e_i, \vec \theta_i, \vec \tau_i \}_{i=1}^N$ randomly
    %\Ensure
    \Repeat
        \State Sample representer points $\{\vec s_c\}_{c=1}^C$ (Section \ref{sec:aces})
        \\
        \State Learn reward model $p(R  \vert \mathcal{D}, \vec s, \vec \theta)$ from $\mathcal{D}$ (Eq.~\ref{eq:gp-model})
        \State Select $(\vec s_q, \vec \theta_q) = \argmax_{\vec s, \vec \theta} \text{ACES}(\vec s, \vec \theta)$  (Eq.~\ref{eq:aces1})
        \State Execute rollout $\vec \tau \sim p(\vec \tau \vert \vec s_q, \vec \theta_q)$ with the robot
        \State Add $(\vec s_q, \vec \theta_q , R(\vec s_q, \vec \tau))$ to $\mathcal{D}$
    \Until{Policy $\pi$ converges}
    \end{algorithmic}
    \end{algorithm}
     \end{minipage}
    \begin{minipage}[t]{.49\textwidth}
    \vspace{0pt}  
    \begin{algorithm}[H]
    \caption{FACES (ours)}
    \label{alg:faces}
    \begin{algorithmic}
    % \Require Initialize dataset $\mathcal{D} = \{\vec s^e_i, \vec \theta_i, \vec \tau_i \}_{i=1}^N$ randomly
    %\Ensure
    \Repeat
        \State Sample representer points $\{\vec s_c\}_{c=1}^C$, $\vec s_c = (\vec s^t_c, \vec s^e_c)$
        \State Construct datasets $\{\mathcal{D}_c \}_{c=1}^C$ from $\mathcal{D}$ (Eq.~\ref{eq:map-data})
        \State Learn reward models $\{\textrm{GP}_c \}_{c=1}^C$ from $\{\mathcal{D}_c \}_{c=1}^C$ (Eq.~\ref{eq:gp-model})
        \State Select $(\vec s^e_q, \vec \theta_q) = \argmax_{\vec s^e, \vec \theta} \text{FACES}(\vec s^e, \vec \theta)$  (Eq.~\ref{eq:aces2})
        \State Execute rollout $\vec \tau \sim p(\vec \tau \vert \vec s^e_q, \vec \theta_q)$ with the robot
        \State Add $(\vec s^e_q, \vec \theta_q , \vec \tau)$ to $\mathcal{D}$
    \Until{Policy $\pi$ converges}
    \end{algorithmic}
    \end{algorithm}
     \end{minipage}
\end{figure*}

\subsection{Bayesian Optimization for Factored CPS} \label{sec:bo-fcps}

Context factorization can be easily incorporated into BO-CPS. The resulting algorithm, Bayesian optimization for factored contextual policy search (BO-FCPS), is shown in Alg.~\ref{alg:bo-fcps}. It maintains a dataset $\mathcal{D} = \{\vec s^e_i, \vec \theta_i, \vec \tau_i \}_{i=1}^N$\footnote{Instead of storing the entire trajectory, in practice we may only record its sufficient statistics for computing the reward, i.e. an outcome $\vec o = \phi(\vec \tau)$.} that can be used to re-evaluate past experiences for a new query context $\vec s_q = (\vec s^t_q, \vec s^e_q)$. Given reward function $R(\vec s^t, \vec \tau)$, we construct a query-specific dataset,
\begin{equation} \label{eq:map-data}
\mathcal{D}_q = \{\vec s^e_i, \vec \theta_i, R(\vec s^t_q, \vec \tau_i) \}_{i=1}^N,
\end{equation}
for learning a specialized reward model,
\begin{equation} \label{eq:gp-model}
p(R^t_q \vert \mathcal{D}_q, \vec s^e_q, \vec \theta) = \mathcal{N}(R^t_q \vert \mu_{\vec s_q, \vec \theta}, \sigma^2_{\vec s_q, \vec \theta}),
\end{equation}
before each rollout. This model is specific to the current target context $\vec s^t_q$. Jointly, the set of all possible reward models $\{p(R^t_{q'} \vert \mathcal{D}_{q'}, \vec s^e, \vec \theta)\}$ w.r.t. arbitrary targets $\vec s^t_{q'}$ generalizes directly over the target context space. Thus, each reward model only needs to generalize over environment contexts $\vec s^e$ and policy parameters $\vec \theta$, leading to a reduced input space compared to the original reward model. This has the added benefit of a smaller search space during optimization. The parameters $\vec \theta$ for context $\vec s_q$ are found by optimizing the acquisition function given the target-specific reward model (Eq.~\ref{eq:gp-ucb},~\ref{eq:bocps-policy}). We employ the DIRECT \cite{jones1993lipschitzian} algorithm for optimization, followed by \mbox{L-BFGS} \cite{byrd1995limited} to refine the result.

We can formally compare BO-FCPS and BO-CPS if we assume that both approaches share the same dataset $\mathcal{D} = \{\vec s_i, \vec \theta_i, \vec \tau_i, \vec R(\vec s^e_i, \vec \tau_i) \}_{i=1}^N$ and the same GP hyper-parameters. In this case, the reward model of BO-FCPS evaluated at a particular target context $\vec s^t_q$ is always at least as accurate as the one learned by BO-CPS. To see this, recall that BO-FCPS differs from BO-CPS in two ways: (1) BO-FCPS re-computes the rewards for $\vec s^t_q$, and (2) BO-FCPS does not consider the rewards at other target contexts $\vec s^t_{q'} \neq \vec s^t_q$. Re-computing the rewards is trivially beneficial because BO-FCPS knows the true reward for target context $\vec s^t_q$ given a trajectory $\vec \tau$, whereas BO-CPS needs to infer the reward from correlations between target contexts. Disregarding rewards at other target contexts $\vec s^t_{q'} \neq \vec s^t_q$ does not degrade the predictive performance of our model either. That is because for a given context-parameter pair the only source of uncertainty w.r.t. their expected reward is in the execution of the trajectory $\vec \tau$ and its effect on the environment. Since the trajectory does not depend on the target context, evaluating the same trajectory under different target contexts does not reveal more information about the trajectory itself. % In other words, a trajectory $\vec \tau$ is conditionally independent of the target context, i.e. $p(\vec \tau \vert \vec s, \vec \theta) = p(\vec \tau \vert \vec s^e, \vec \theta)$. % that determines its quality, i.e. $p(\vec \tau \vert \vec s, \vec \theta) = p(\vec \tau \vert \vec s^e, \vec \theta)$. Indeed, we found empirically that re-computing the rewards for more target contexts and learning a reward model from a joint dataset $\bigcup_{\vec s^t \neq \vec s^t_q} \{\vec s^e_i, \vec \theta_i, R(\vec s^t, \vec \tau_i) \}_{i=1}^N~\cup~\mathcal{D}_q$ does not lead to better reward predictions. 

Note that while a better reward model at a given target context leads to better greedy performance, e.g. during offline evaluation, it does not necessarily imply higher cumulative rewards during learning. Furthermore, in practice both the dataset $\mathcal{D}$ and the GP hyper-parameters would be different. Empirically, BO-FCPS does achieve both higher online and offline performance, as we show in Section~\ref{sec:experiments}. We defer a more extensive theoretical analysis to future work.
% Note that in practice both the dataset $\mathcal{D}$ and the GP hyper-parameters would be different. A better reward model at a given target context leads to better greedy performance, e.g. during offline evaluation, however, it does not necessarily imply higher rewards during learning. In this paper we empirically evaluate online and offline performance, and defer a more extensive theoretical analysis to future work.

\begin{figure*}[t]
    \centering
    \begin{subfigure}[b]{0.32\textwidth}
        \includegraphics[width=\textwidth]{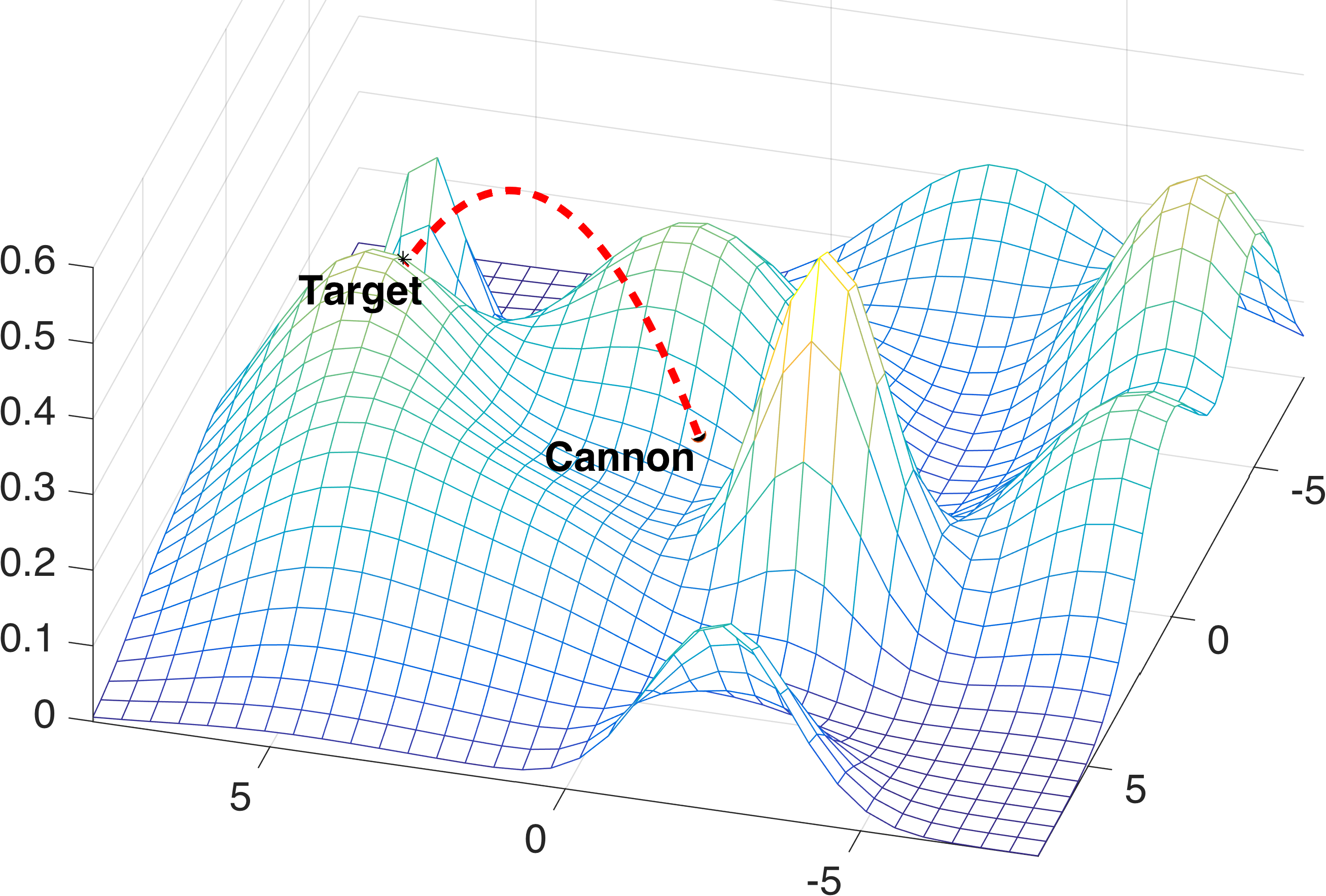}
        \caption{Toy cannon setup}
        \label{fig:toy-setup}
    \end{subfigure}
    \begin{subfigure}[b]{0.32\textwidth}
        \includegraphics[width=\textwidth]{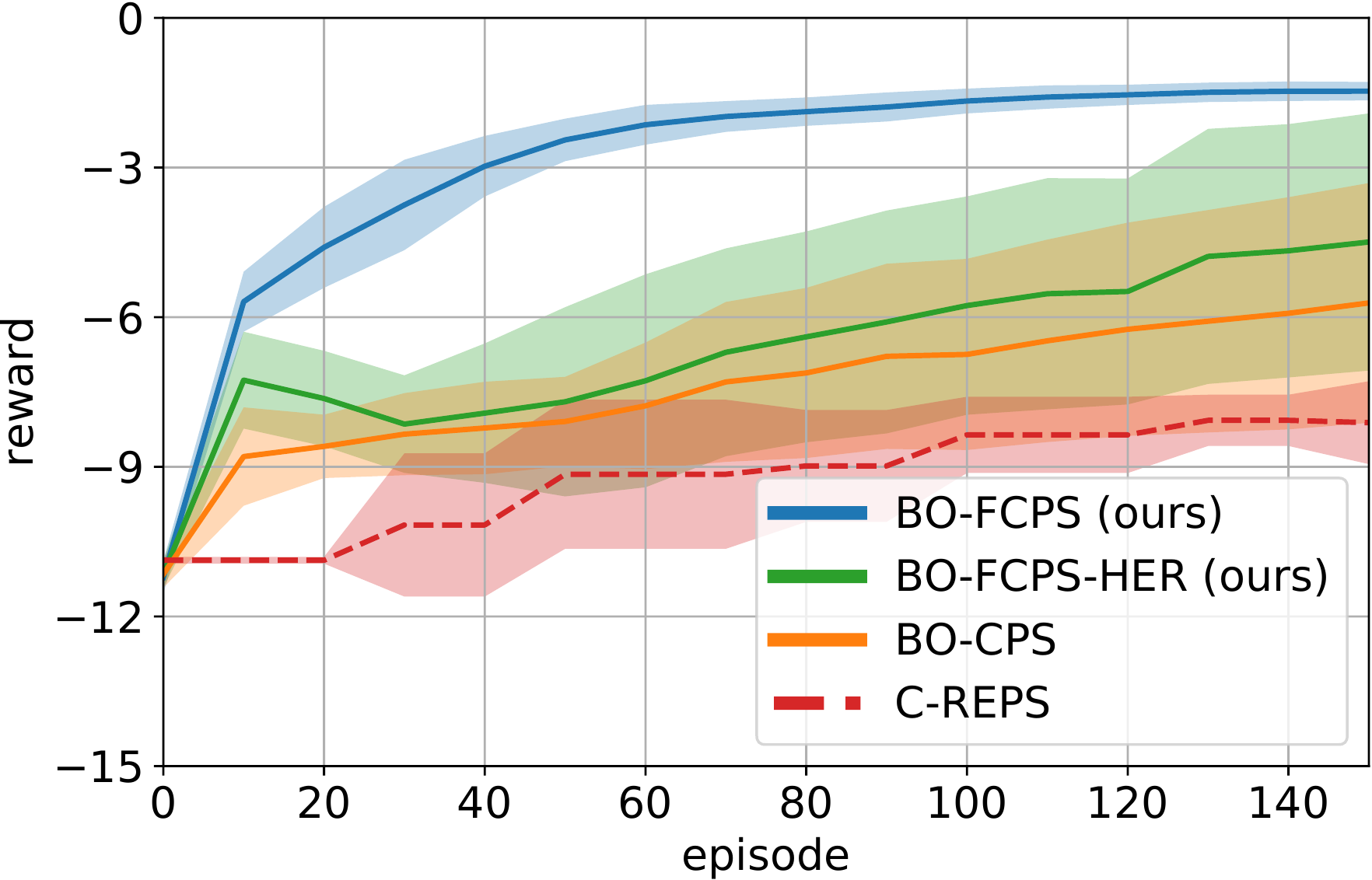}
        \caption{Comparison of different algorithms}
        \label{fig:toy-performance}
    \end{subfigure}
    \begin{subfigure}[b]{0.32\textwidth}
        \includegraphics[width=\textwidth]{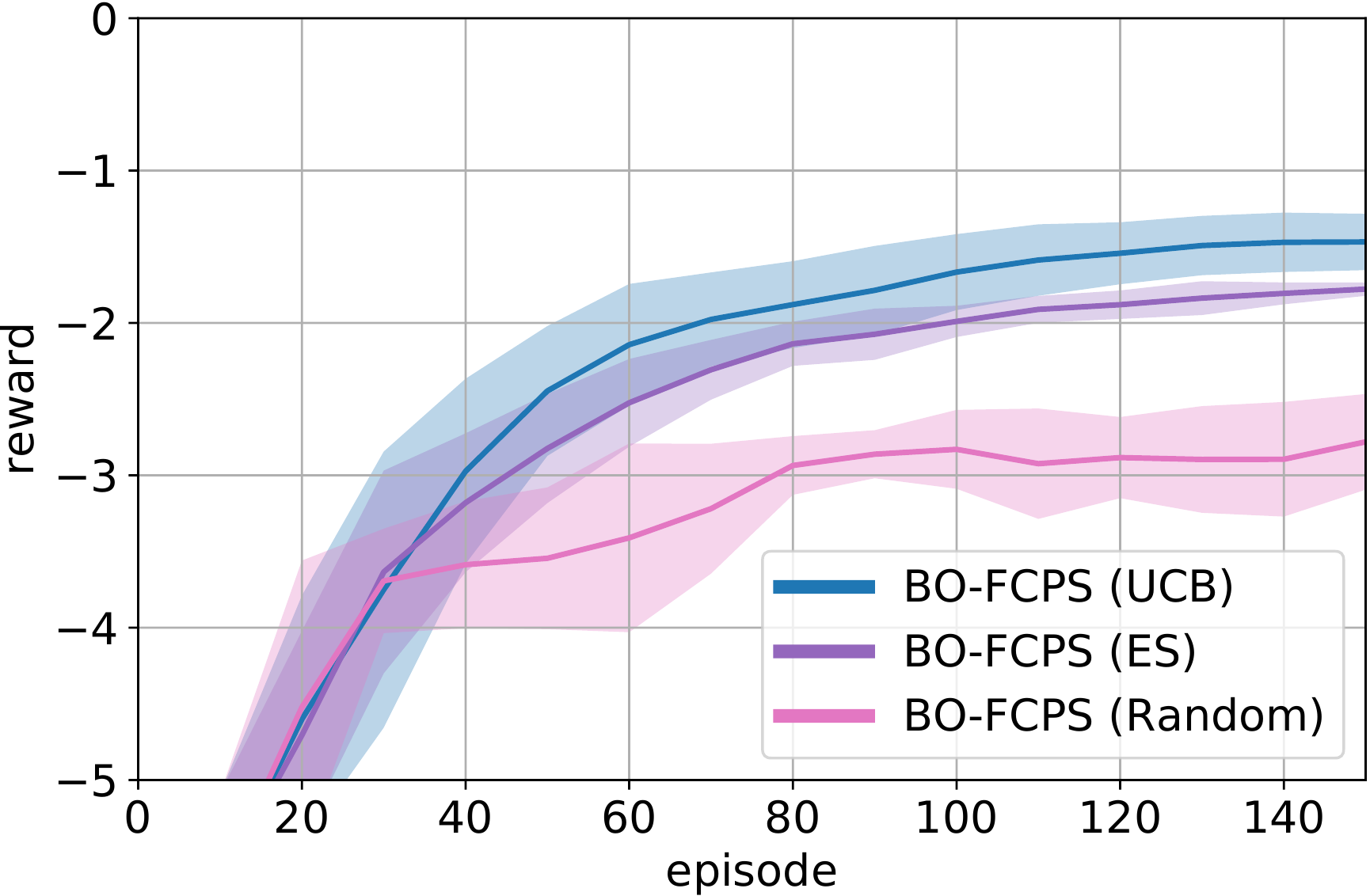}
        \caption{Comparison of different variants}
        \label{fig:toy-variants}
    \end{subfigure}
    \caption{\textbf{(a)} Visualization of the toy cannon task. \textbf{(b)-(c)} Offline performance evaluated on a fixed set of contexts from a $15 \times 15$ grid. Results are averaged over 10 randomly generated environments. Shaded areas denote one standard deviation.
    }
\end{figure*}

\subsection{Factored Active Contextual Entropy Search} \label{sec:faces}
In the active learning setting, the agent chooses both the policy parameters $\vec \theta$ and the context $\vec s$. Applying GP-UCB is problematic as it would not take the varying difficulty of tasks into account~\cite{fabisch2014active}. % Furthermore, the reward distributions become non-stationary, violating the stationarity assumption of UCB.
Instead, we follow ACES~\cite{metzen2015active} and use an ES-based acquisition function, which aims to choose the most informative query points for global optimization~\cite{hennig2012entropy}. 

We integrate factored contexts into ACES as follows. In each iteration, we map previous experience to all representer points $\{\vec s_c\}_{c=1}^C$ in the context space, i.e. we construct $C$ different datasets $\{\mathcal{D}_c \}_{c=1}^C$ as in Eq.~\ref{eq:map-data}. From these datasets, we construct a set of GP models $\{\textrm{GP}_c \}_{c=1}^C$ that we use to evaluate the ACES acquisition function. In particular, we employ the corresponding $\textrm{GP}_c: p(R_c^t|\mathcal{D}_c, \vec s^e_c, \vec \theta)$ when the expected information gain $G^{\vec s}(\vec s_q, \vec \theta_q)$ after a hypothetical query $(\vec s_q, \vec \theta_q)$ is evaluated for $\vec s = \vec s_c$. Similar to BO-FCPS, we therefore directly use the target-specific GPs instead of relying on the correlation between target contexts.

Note that the choice on the target-type query context $\vec s_q^t$ is actually indifferent if we ignore rewards during training, and thus we only need to select $(\vec s_q^e, \vec \theta_q)$ by maximizing
\begin{equation} \label{eq:aces2}
\text{FACES}(\vec s^e_q, \vec \theta_q) = \sum\nolimits_{c=1}^C G^{\vec s_c}(\vec s^e_q, \vec \theta_q).
\end{equation}
We call the resulting algorithm factored active contextual entropy search (FACES). The algorithm is shown in Alg.~\ref{alg:faces}.

\subsection{Hindsight Experience Replay for Factored CPS} \label{sec:her}

So far we have assumed that the agent has access to the reward function $R(\vec s^t, \vec \tau)$. If we cannot query the reward function at arbitrary points, it is still possible to leverage factored contexts. In particular, we replace the current target context $\vec s^t_q$ after a rollout by the achieved target $ \vec s^t_{\vec \tau}$ and evaluate it again, yielding reward $R(\vec s^t_{\vec \tau}, \vec \tau)$. Thus, the only requirement is to be able to obtain the sample reward $R(\vec s^t_{\vec \tau}, \vec \tau)$ in addition to the actual reward $R(\vec s^t_q, \vec \tau)$. The additional data point $(\vec s^e_q, \vec \theta_q, R(\vec s^t_{\vec \tau}, \vec \tau))$ can then be added to the training dataset $\mathcal{D}$, and standard CPS methods such as BO-CPS, cost-regularized kernel regression \cite{kober2012reinforcement} or contextual relative entropy policy search (C-REPS) \cite{daniel2012hierarchical, kupcsik2013data} can be used without further modifications. Such an approach can be seen as an extension of HER to the CPS setting. We believe we are the first ones that explicitly make this connection.

\section{EXPERIMENTS AND RESULTS} \label{sec:experiments}

We perform experiments to answer the following questions: (a) does context factorization lead to more data-efficient learning for passive and active BO-based CPS algorithms; (b) how does the choice of acquisition function influence the performance; (c) does context factorization improve generalization; (d) is our method effective in more complex robotic domains? We address questions (a)-(c) through experiments on a toy cannon task, and (d) on three simulated tasks from the OpenAI Gym~\cite{brockman2016openai}. For the Gym tasks, we employ an extension~\cite{kober2010movement} of the DMP framework~\cite{ijspeert2003learning} to efficiently generate goal-directed trajectories.

\subsection{Toy Cannon Task}

The toy cannon task is a popular domain for evaluating CPS algorithms~\cite{da2014active, metzen2015bayesian, metzen2015active}. As shown in Fig.~\ref{fig:toy-setup}, a cannon is placed in the center of a 3D coordinate system and has to shoot at targets on the ground in the range of $[-11, 11] \times [-11, 11]$m. The contextual policy maps from 2D targets $\vec s^t \in \mathcal{R}^2$ to 3D launch parameters $\vec \theta \in \mathcal{R}^3$: horizontal orientation $\alpha \in [0, 2\pi]$, vertical angle $\beta \in [0.01, \pi/2 - 0.2]$ and speed $v \in [0.1, 5]$ ms. The reward function is given by $R(\vec s^t, \vec \tau) = -\|\vec s^t - \vec s^t_{\vec \tau} \| - 0.05 v^2$, where $\vec s^t_{\vec \tau}$ is the achieved hitting location of a trajectory $\vec \tau$. To increase the difficulty of the problem, we add Gaussian noise ($\sigma_n = 1^{\circ}$) to the desired launch angle during training and randomly place hills in the environment. The learning agent is unaware of the hills and the target contexts carry no information on the elevation.

\begin{table}[t]
\centering
\begin{tabular}{@{}lcll@{}}
\toprule
\multicolumn{1}{c}{} & $t=50$ & $t=100$ & $t=150$ \\ \midrule
C-REPS & -496 ($\pm$ 17) & -955 ($\pm$ 56) & -1357 ($\pm$ 62) \\
BO-CPS & -461 ($\pm$ 28) & -843 ($\pm$ 70) & -1148 ($\pm$ 151) \\
BO-FCPS-HER (ours) & -447 ($\pm$ 24) & -809 ($\pm$ 71) & -1111 ($\pm$ 140) \\
BO-FCPS (ours) & \textbf{-303 ($\pm$ 34)} & \textbf{-414 ($\pm$ 44)} & ~\textbf{-499 ($\pm$ 56)} \\ \bottomrule
\end{tabular}
\caption{Online learning performance on the toy cannon task averaged over 10 random seeds. We report mean cumulative rewards obtained during the first $t$ iterations.}
\label{tab:toy-rewards}
\end{table}

\begin{figure}[t]
    \centering
    \begin{subfigure}[b]{0.23\textwidth}
        \includegraphics[width=\textwidth]{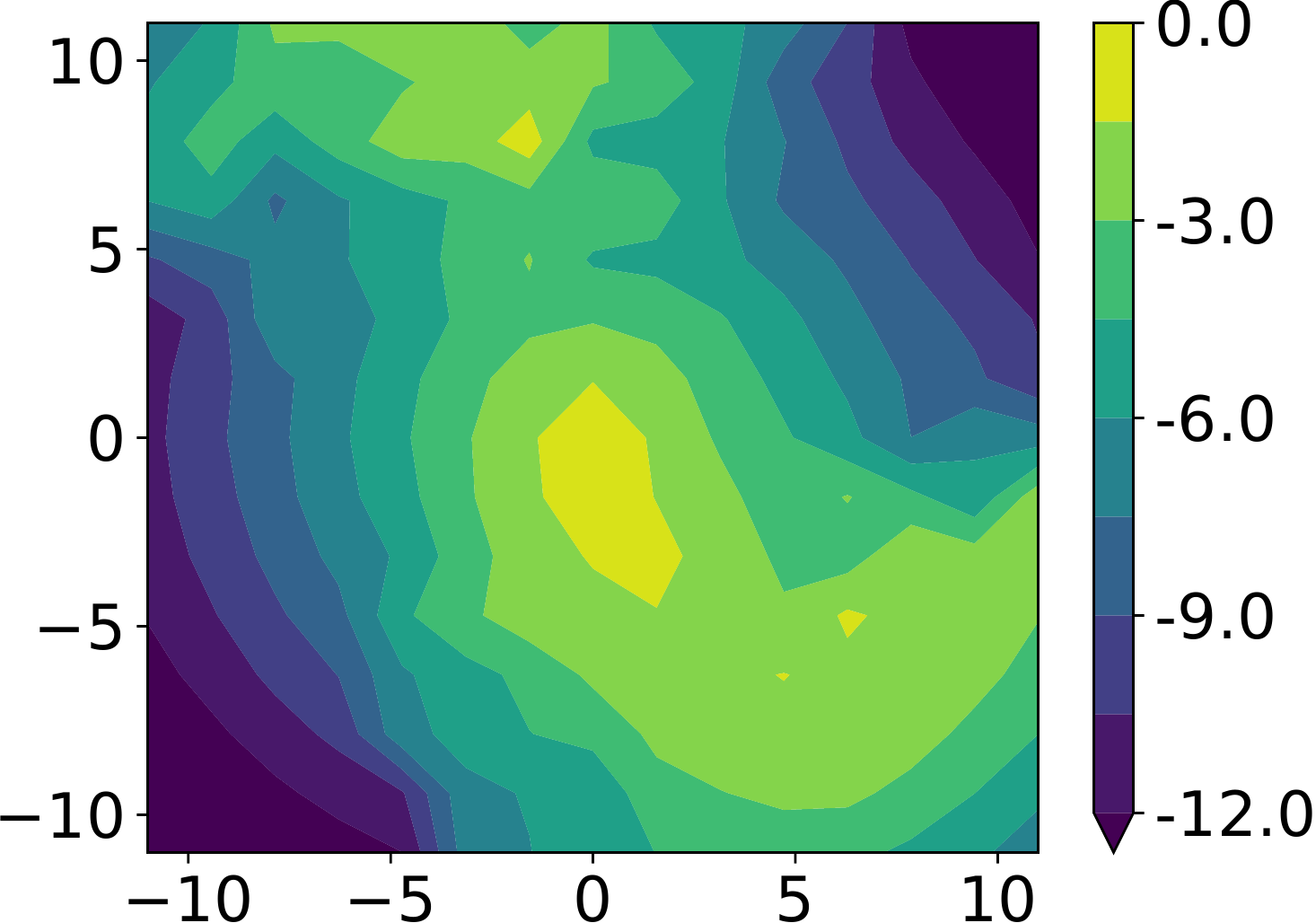}
        \caption{BO-CPS}
    \end{subfigure}
    \begin{subfigure}[b]{0.23\textwidth}
      \includegraphics[width=\textwidth]{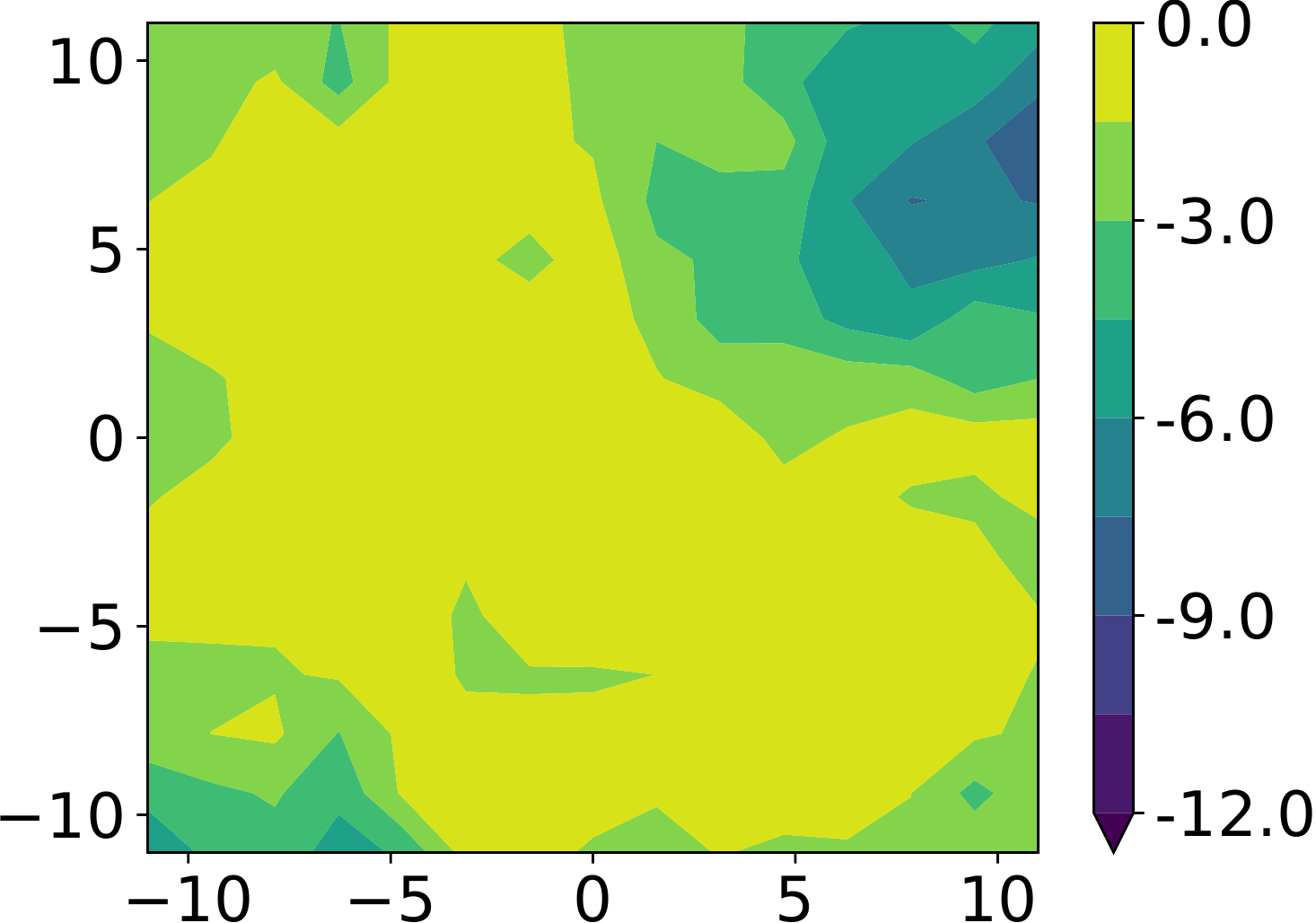}
        \caption{BO-FCPS (ours)}
    \end{subfigure}
    \caption{Achieved rewards in target context space after 150 episodes, where no contexts were sampled from the upper-right and lower-left corner during training. %Contour lines denote reward levels. The low-reward region of BO-FCPS is due to a hill in the landscape.
    \vspace{-1.5em}
    }
    \label{fig:toy-rewards}
\end{figure}

First, we compare both our factored BO approach (BO-FCPS) and our factored HER-style BO approach (BO-FCPS-HER) to standard BO-CPS. Each algorithm uses the GP-UCB acquisition function. We employ a zero-mean GP prior, $p(\vec f) \sim \text{GP}(\vec 0, k(\vec x, \vec x'))$, with squared-exponential kernel $k(\vec x, \vec x') = \sigma_f^2 \exp\left(-\frac{1}{2} (\vec x - \vec x')\tran \vec \Lambda^{-1} (\vec x - \vec x')\right)$, where $\vec x = (\vec s, \vec \theta)$, and $\vec \Lambda = \diag(\vec \ell)$ contains positive length-scale parameters $\vec \ell$. The GP hyper-parameters are optimized by maximizing the marginal likelihood. We also compare to C-REPS, which performs local policy updates instead of global optimization. We use a linear Gaussian policy with squared context features that is updated every 30 episodes subject to the relative entropy bound $\epsilon = 0.5$.

The offline performance of each algorithm is shown in Fig.~\ref{fig:toy-performance}. BO-FCPS requires only 60 episodes to find a good policy, a considerable improvement over standard BO-CPS.  BO-FCPS-HER improves on BO-CPS as well, although the variance is much larger. This is because the GP hyper-parameter optimization sometimes got stuck in a local minimum, overfitting to the context variables while ignoring the influence of the parameters $\vec \theta$. We hypothesize that a full Bayesian treatment would mitigate this issue. C-REPS is not competitive on this low-dimensional task since the policy adapts too slowly. The above findings are confirmed by the online performances, which are summarized in Table~\ref{tab:toy-rewards}.

In Fig.~\ref{fig:toy-variants}, we evaluate the dependence of BO-FCPS on the acquisition function. We compare three variants: BO-FCPS with GP-UCB~(UCB), entropy search~(ES) and a random acquisition function. Using GP-UCB over ES leads to slightly faster learning as ES tends to explore too much. Likewise, random exploration is not sufficient for data-efficient learning. We therefore focus on GP-UCB.

\begin{figure}[t]
    \centering
    \includegraphics[width=.32\textwidth]{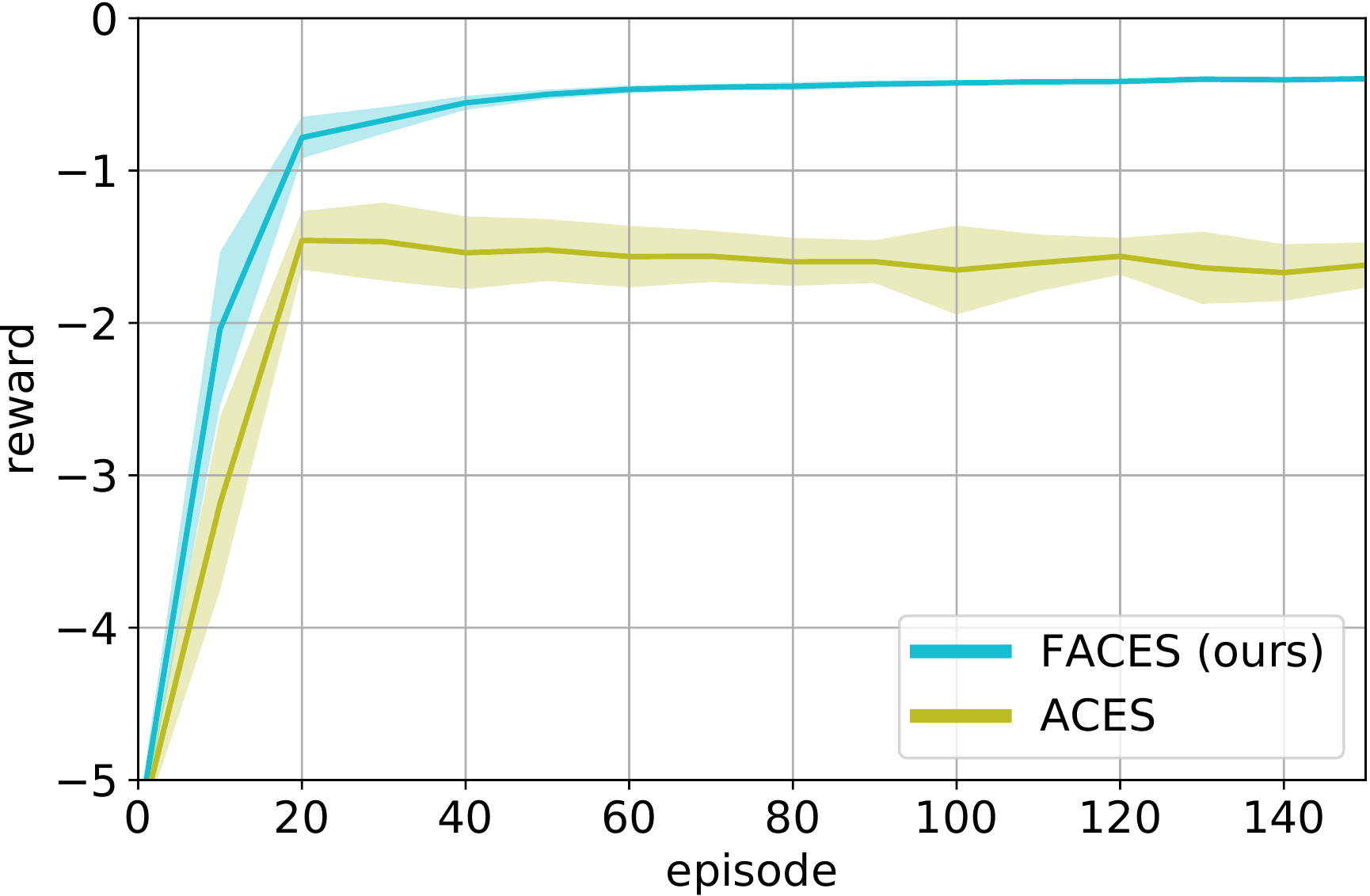}
    \caption{Learning curve for active learning setting averaged over 10 randomly generated environments, evaluated offline based on contexts placed uniformly on an $8 \times 8$ grid. Shaded areas denote one standard deviation.
    \vspace{-1.5em}
    }
    \label{fig:toy-active}
\end{figure}

\begin{figure*}[t]
    \vspace{1em}
    \centering
    ~~
    \begin{subfigure}[b]{0.2\textwidth}
        \includegraphics[width=\textwidth]{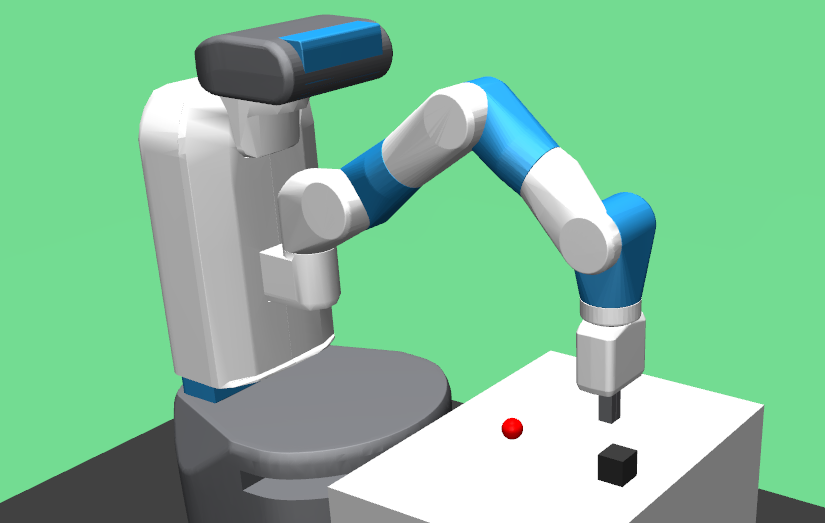}
        %\caption{FetchPush-v1}
    \end{subfigure}
    \qquad\qquad\quad~
    \begin{subfigure}[b]{0.2\textwidth}
        \includegraphics[width=\textwidth]{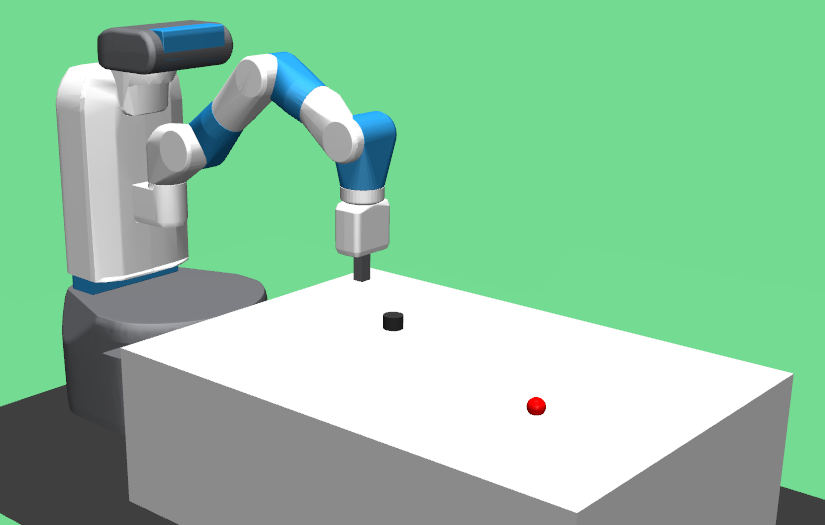}
        %\caption{FetchSlide-v1}
    \end{subfigure}
    \qquad\qquad\quad~
    \begin{subfigure}[b]{0.2\textwidth}
        \includegraphics[width=\textwidth]{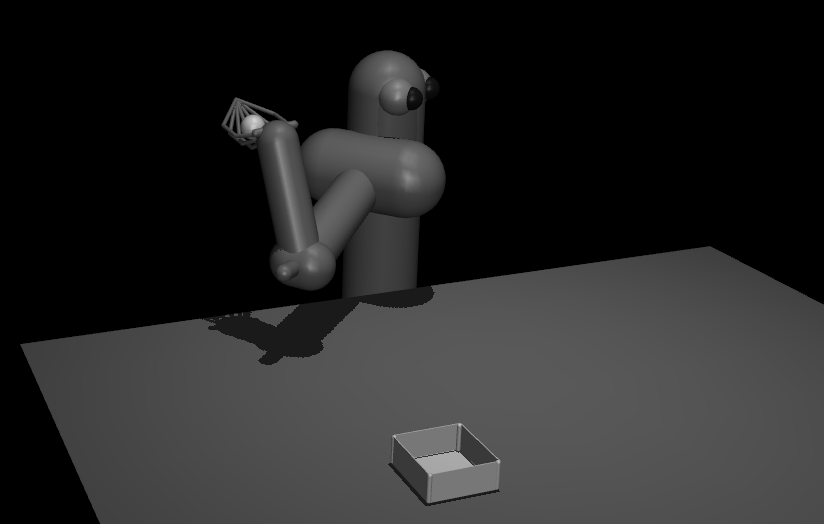}
        %\caption{Thrower-v2}
    \end{subfigure}
    \\[.5em]
    \begin{subfigure}[b]{0.32\textwidth}
        \includegraphics[width=\textwidth]{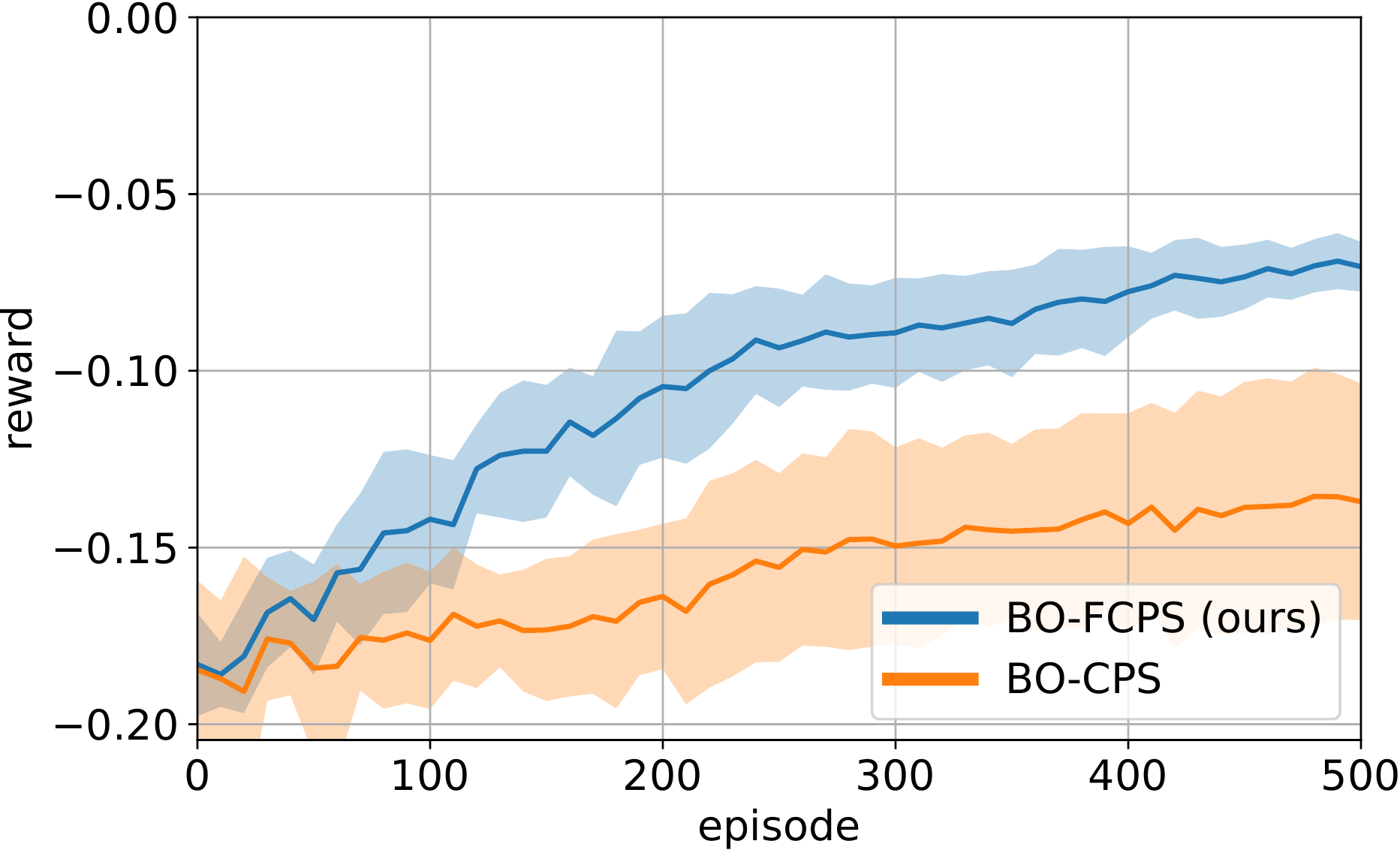}
        \caption{FetchPush-v1}
    \end{subfigure}
    \begin{subfigure}[b]{0.32\textwidth}
        \includegraphics[width=\textwidth]{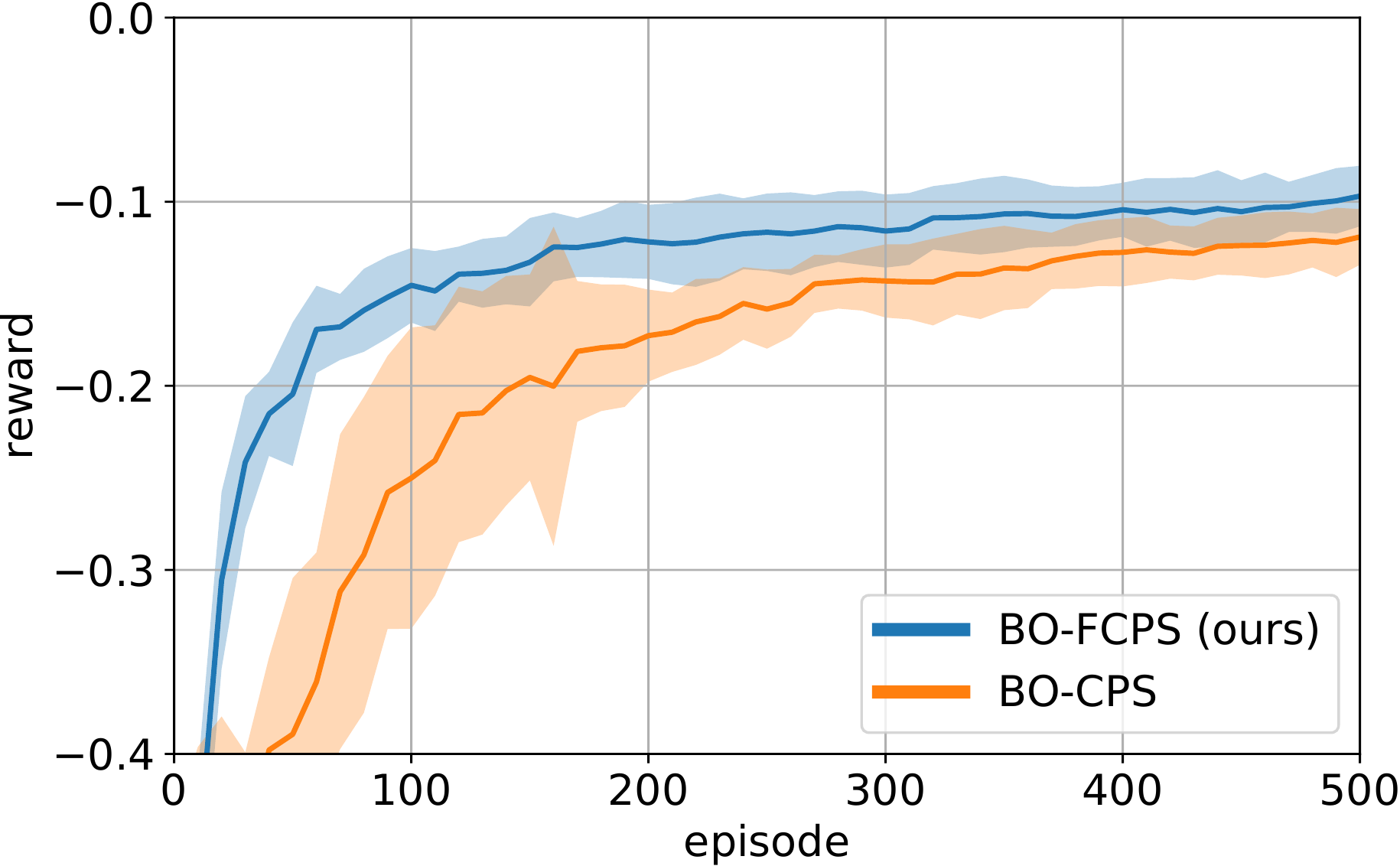}
        \caption{FetchSlide-v1}
    \end{subfigure}
    \begin{subfigure}[b]{0.32\textwidth}
        \includegraphics[width=\textwidth]{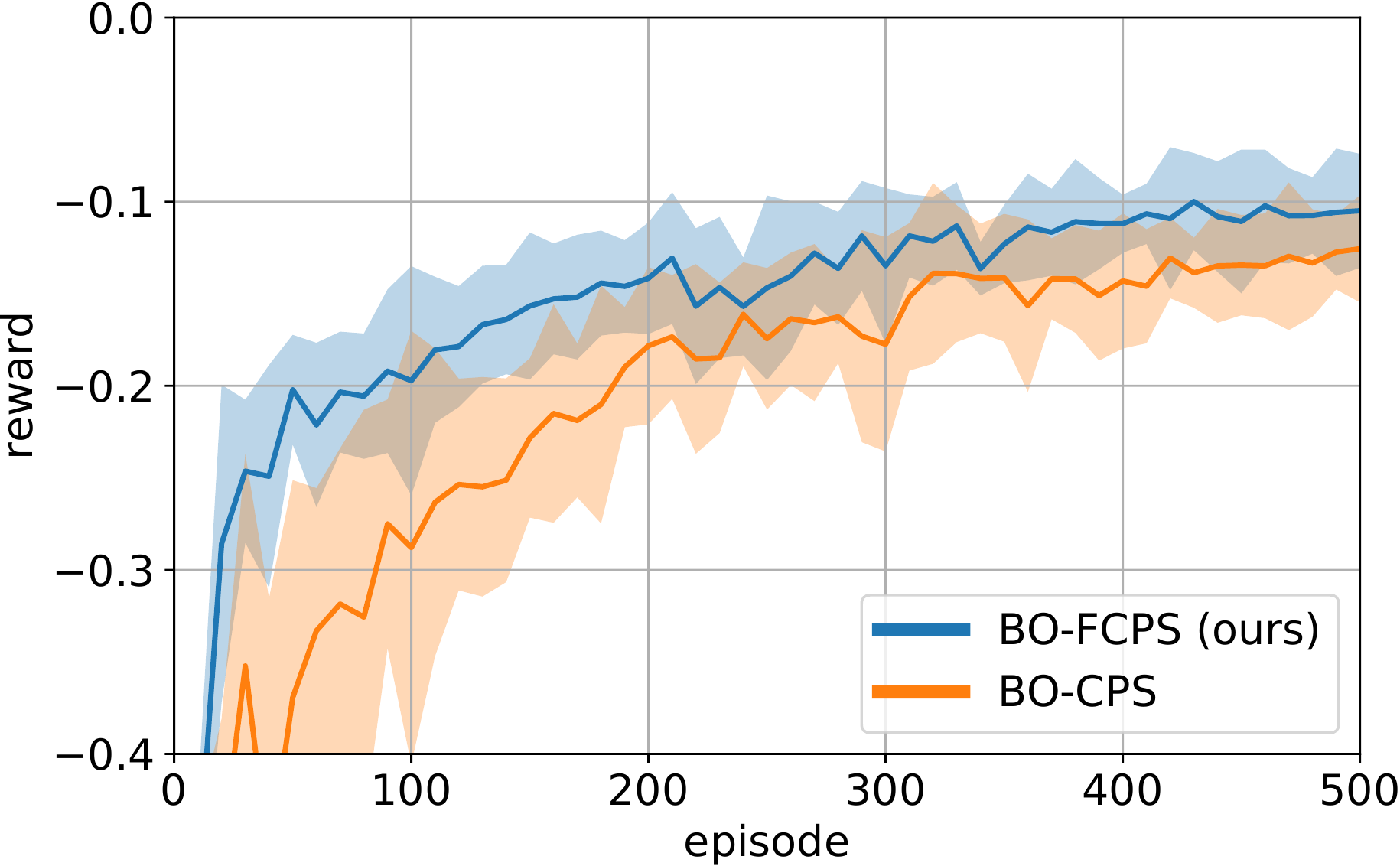}
        \caption{Thrower-v2}
    \end{subfigure}
    \caption{Learning curves for the OpenAI Gym tasks. Policies are evaluated after every 10 rollouts, in 25 fixed contexts sampled once before the experiment. Results are averaged over 10 random seeds. Shaded areas denote one standard deviation. \vspace{-0.5em}}
    \label{fig:sim-performance}
\end{figure*}

Next, we demonstrate why evaluating previous rollouts for the given target context leads to improved generalization. We only present contexts $\vec s^t \in [-11, 0] \times [0, 11]~\cup~[0, 11] \times [-11, 0]$ during training, while the agent has to generalize to the entire context space during evaluation. As depicted in Fig.~\ref{fig:toy-rewards}, BO-FCPS generalizes much better to unseen contexts due to a much more accurate reward model at locations where it has already shot to. When evaluated in previously unseen contexts, the mean rewards achieved by BO-FCPS were higher by $4.0$; while the improvement was $3.05$ in contexts that were sampled during training. 
% Shooting into the upper-right or lower-left corner thus results in low reward during training (since no target contexts were sampled in those regions) but the same launch parameters may achieve high reward when another target is in the shooting direction during evaluation.

Finally, we consider an active learning setting, where the agent observes an additional context variable $\mathbb{I} \in [0, 1]$ that indicates whether the learning agent should shoot or not. If $\mathbb{I} \leq 0.1$, the agent receives reward $R(\vec s^t, \vec \tau) = -\|\vec s^t - \vec s^t_{\vec \tau} \| - 0.05 v^2$ as before, and an action penalty $-\|\vec \theta \|$ otherwise. The agent should therefore actively select $\mathbb{I} \leq 0.1$, for which the reward function is much harder to learn. We compare FACES to ACES, and use 200 representer points to approximate the acquisition functions. The results are shown in Fig.~\ref{fig:toy-active}. Similar to the passive case, factorization is greatly beneficial: FACES achieves much faster learning than ACES.

\subsection{Simulated Robotic Tasks}

Finally, we apply BO-FCPS to three distinct robotics tasks from the \mbox{OpenAI} Gym \cite{brockman2016openai, plappert2018multi}, namely:
\begin{itemize}
    \item \textbf{FetchPush-v1}: Push a box to a goal position.
    \item \textbf{FetchSlide-v1}: Slide a puck to a goal position that is out of reach of the robot.
    \item \textbf{Thrower-v2}: Throw a ball to a goal position.
\end{itemize}
The BO-FCPS algorithm is used to select the DMP parameters for performing each task according to the current context. We deviate from the original task specification of Thrower-v2 by replacing the joint-space controller in favor of task-space control to reduce the dimensionality of the problem. The same controller is employed in the Fetch environments. Moreover, we use the final distance between the object and the target context as the reward function in each environment. For more details, please refer to~\cite{brockman2016openai, plappert2018multi}.

For FetchPush and FetchSlide, both the initial object position $\vec s^e \in \mathcal{R}^2$ and the desired goal position $\vec s^t \in \mathcal{R}^2$ are varied. The lower-level policy consists of two 3-dimensional task-space DMPs that are sequenced together. The first DMP is used to bring the robot arm into position to manipulate the object, where the trajectory is modulated by 25 basis functions per dimension, and the shape parameters $\vec w$ are learned by imitation. The second DMP is used to execute the actual movement (i.e. pushing, sliding), starting from the final position of the first DMP. The upper-level policy adapts the approach angle $\alpha$ of the first DMP w.r.t. the object, yielding a goal position $\vec y_1$ that is a fixed distance away from the object, and the goal position $\vec y_2$ of the second DMP. We use $\alpha \in [0, \pi], \vec y_2 \in [0, 0.4] \times [-0.4, 0.4]$ for FetchSlide and $\alpha \in [0, 2\pi], \vec y_2 \in [-0.2, 0.2]^2$ for FetchPush.

For the Thrower environment, only the desired goal position $\vec s_t \in \mathcal{R}^2$ is varied. The lower-level policy is a single 3-dimensional task-space DMP with 25 basis functions, where the shape parameters are learned by imitation. The upper-level policy selects the goal position $\vec y \in [-0.5, 0.5] \times [1, 1.5] \times [-0.5, 0.5]$ and goal velocity $\dot{\vec y} \in [0, 1]^3$ of the DMP, resulting in a 6-dimensional parameter vector $\vec \theta$.

Results are shown in Fig.~\ref{fig:sim-performance}. Our proposed approach consistently outperforms standard BO-CPS, suggesting that our earlier findings on the toy cannon task apply to more complex simulated robotic domains as well. 

\section{DISCUSSION AND FUTURE WORK}

We introduced context factorization and integrated it into passive and active learning approaches for CPS with BO. The improvement we can expect from factorization depends on the characteristics of the task. It is most effective if a large part of the learning challenge is about generalizing across contexts, as opposed to learning a good policy for a single context. In general, the larger the space of target contexts over environment contexts and policy parameters, the more factorization is expected to be beneficial.

In this paper, we focused on BO for CPS. One shortcoming of BO is that it does not scale well to high-dimensional problems. Future work may address scalability and explore alternative acquisition functions such as predictive entropy search~\cite{hernandez2014predictive}. Since context factorization is not specific to BO, it could also be applied to other CPS algorithms, e.g. C-REPS~\cite{kupcsik2013data} or contextual CMA-ES~\cite{abdolmaleki2017contextual}. A simple approach would be to populate the dataset with additional re-evaluated samples, similar to HER. Finally, we demonstrated the benefits of factorization in extensive simulated experiments. In the future, we plan to demonstrate the approach on a real robot system as well.

\addtolength{\textheight}{-8cm}   % This command serves to balance the column lengths

%%%%%%%%%%%%%%%%%%%%%%%%%%%%%%%%%%%%%%%%%%%%%%%%%%%%%%%%%%%%%%%%%%%%%%%%%%%%%%%%

%%%%%%%%%%%%%%%%%%%%%%%%%%%%%%%%%%%%%%%%%%%%%%%%%%%%%%%%%%%%%%%%%%%%%%%%%%%%%%%%

%%%%%%%%%%%%%%%%%%%%%%%%%%%%%%%%%%%%%%%%%%%%%%%%%%%%%%%%%%%%%%%%%%%%%%%%%%%%%%%%

%\section*{APPENDIX}
%\subsection{Experimental Setup}

% \section*{ACKNOWLEDGMENT}
% Put sponsor acknowledgments in the unnumbered footnote on the first page.

%%%%%%%%%%%%%%%%%%%%%%%%%%%%%%%%%%%%%%%%%%%%%%%%%%%%%%%%%%%%%%%%%%%%%%%%%%%%%%%%
\newpage
\bibliography{paper} 
\bibliographystyle{ieeetr}

\end{document}